\documentclass[conference]{IEEEtran}
\IEEEoverridecommandlockouts
\pdfoutput=1
\usepackage[font=small,labelfont=bf]{caption}
\usepackage{cite}
\usepackage{amsmath,amssymb,amsfonts}
\usepackage{algorithmic}
\usepackage{graphicx}
\usepackage{textcomp}
\usepackage{subcaption}
\usepackage{hyperref}
\usepackage{wrapfig}
\def\BibTeX{{\rm B\kern-.05em{\sc i\kern-.025em b}\kern-.08em
    T\kern-.1667em\lower.7ex\hbox{E}\kern-.125emX}}
\begin{document}

\title{Hierarchical internal representation of spectral features in deep convolutional networks\\trained for EEG decoding}

\author{\IEEEauthorblockN{1\textsuperscript{st} Kay Gregor Hartmann}
\IEEEauthorblockA{\textit{Translational Neurotechnology Lab} \\
\textit{Medical Center - University of Freiburg}\\
Freiburg, Germany \\
hartmank@tf.uni-freiburg.de}
\and
\IEEEauthorblockN{2\textsuperscript{nd} Robin Tibor Schirrmeister}
\IEEEauthorblockA{\textit{Translational Neurotechnology Lab} \\
\textit{Medical Center - University of Freiburg}\\
Freiburg, Germany \\
robin.schirrmeister@uniklinik-freiburg.de}
\and
\IEEEauthorblockN{3\textsuperscript{rd} Tonio Ball}
\IEEEauthorblockA{\textit{Translational Neurotechnology Lab} \\
\textit{Medical Center - University of Freiburg}\\
Freiburg, Germany \\
tonio.ball@uniklinik-freiburg.de}
}

\maketitle

\begin{abstract}
Recently, there is increasing interest and research on the interpretability of machine learning models, for example how they transform and internally represent EEG signals in Brain-Computer Interface (BCI) applications. This can help to understand the limits of the model and how it may be improved, in addition to possibly provide insight about the data itself.
Schirrmeister et al. (2017) have recently reported promising results for EEG decoding with deep convolutional neural networks (ConvNets) trained in an end-to-end manner and, with a causal visualization approach, showed that they learn to use spectral amplitude changes in the input.
In this study, we investigate how ConvNets represent spectral features through the sequence of  intermediate stages of the network. We show higher sensitivity to EEG phase features at earlier stages and higher sensitivity to EEG amplitude features at later stages. Intriguingly, we observed  a specialization of individual stages of the network to the classical EEG frequency bands alpha, beta, and high gamma. Furthermore, we find first evidence that particularly in the last convolutional layer, the network learns to detect more complex oscillatory patterns beyond spectral phase and amplitude, reminiscent of the representation of complex visual features in later layers of ConvNets in computer vision tasks.
Our findings thus provide  insights into how ConvNets hierarchically represent spectral EEG features in their intermediate layers and suggest that ConvNets can exploit and might help to better understand the compositional structure of EEG time series.
\end{abstract}


\section{Introduction}
Recently, there is increasing interest and research on the interpretability of machine learning models. Interpretability is for example important  in brain signal decoding and brain-computer interfaces (BCIs)\cite{Haufe2014,Blankertz2008,Sturm2016}. In many situations, practitioners in this field want to understand how a machine learning model extracts information from the brain-signal. Aided with such an understanding, practitioners can better understand what brain-signal features a model uses and develop ideas how to improve information extraction. Recent interest in interpretability is also motivated by emerging machine learning applications in safety critical environments, for example a BCI application that lets the user control  a wheelchair. For several well-established EEG decoding approaches, methods have been developed to understand what is learned from the signal\cite{Haufe2014,Blankertz2008,Schirrmeister2017a}.

Lately, deep convolutional neural networks (deep ConvNets) have shown promising results in EEG decoding. ConvNets can exploit the hierarchical structure present in many natural signals\cite{Goodfellow-et-al-2016,Schmidhuber2014,LeCun2015} and have shown groundbreaking results in computer vision and speech recognition\cite{Abdel-Hamid2014,Sainath2013,Krizhevsky2012}. First results suggest that they may perform at least as good as already well-established EEG decoding approaches (with already dedicated visualization methods)\cite{Schirrmeister2017a,Thodoroff2016,Shamwell2016}. Notably, ConvNets can reach good accuracies learning from the raw EEG in an end-to-end manner without any hand-designed feature extraction.

However, deep ConvNets are notoriously hard to interpret. For  visualizing ConvNets trained on EEG, studies used weights and outputs of different layers of the ConvNets, or computed saliency maps that show how small changes in the input would affect the decoding decision\cite{Zeiler2013,Kindermans2017,Bojarski2016,Simonyan2013,Schirrmeister2017a}. Schirrmeister et al. (2017) further developed a visualization method that computes correlations between amplitude changes in different frequency bands of the EEG signal with changes in the resulting final output of the ConvNet, showing that ConvNets do use frequency-specific spectral amplitude. Still, many aspects are not well understood, especially about \textit{how} ConvNets use their multiple computational stages to extract spectral EEG features, both amplitude and phase, in different frequency bands - aspects which might shed further light on how ConvNets perform EEG decoding.

Therefore, here we investigate several basic, yet not fully understood aspects of the internal representation of EEG data in ConvNets trained in an end-to-end manner. For this, we investigated a ConvNet architecture that Schirrmeister et al. (2017) recently showed to reach good accuracies for decoding task-related information from EEG. As we aim to obtain insights about which features are learned at which point of the ConvNet's hierarchical structure, we apply the perturbation method described in Schirrmeister et al. (2017) to the intermediate representations of the ConvNet layers. Furthermore, we extend the method to also investigate how strongly the ConvNet's internal representations react to changes in phase features of the signal. Additionally, we look at the most-activating inputs, i.e., the inputs that lead to the largest activations for specific filters. We investigate if they resemble sinusoidal shapes and also visualize some of the most activating inputs directly. By these analyses, we aim to get a better understanding about how ConvNets learn to extract spectral information from the brain signal.

The findings of the present study shed light on the internal representation of spectral features in ConvNets that allow them to learn frequency-specific spectral features from EEG in an end-to-end manner. We find a consistent pattern that the representations computed by earlier layers change more strongly when the phase is perturbed while the representations of later layers react more strongly to amplitude changes. Interestingly, we find that for each convolution-pooling-block after the first, the internal representation becomes phase-invariant for one of the classical EEG frequency bands at a time, starting with the high gamma over the beta to the alpha band. We also find some preliminary evidence that the later internal representations might represent more complex patterns beyond simple sinusoidal shapes.
\begin{figure*}[h]
    \centering
        \includegraphics[width=0.99\textwidth]{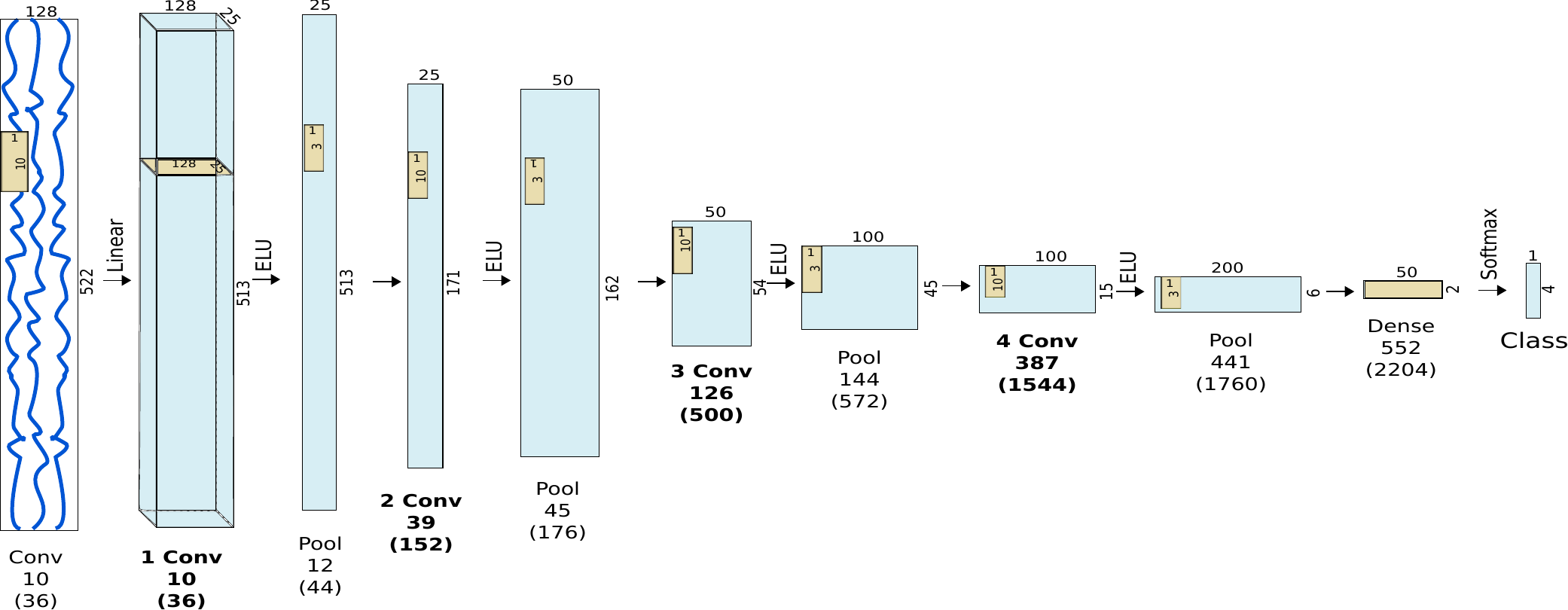}
        \caption{ConvNet architecture for EEG decoding proposed by Schirrmeister et al. (2017). Blue rectangles are layer inputs, yellow rectangles are kernels. Receptive field size of the activation units in a layer are noted below the layer name in samples and (ms). There are 5 convolutional layers in total. The first convolutional layer is not followed by a non-linear activation function, but passes its activations directly to the next convolutional layer (see Schirrmeister et al. 2017 for an explanation for this design choice). The following 4 convolutional layers are each followed by a layer with exponential linear units\cite{Clevert2015} and a max-pooling layer\cite{LeCun2010}. The second convolutional kernel spans all EEG channels, therefore only the time and filter dimensions are left for subsequent layers. We investigated the 4 convolutional layers denoted by 1, 2, 3 and 4.}
        \label{fig:Architecture}
\vspace{-15pt}
\end{figure*}
\section{Materials \& Methods}
\subsection{Dataset and pre-processing}
We use the same dataset as in Schirrmeister et al. (2017) which consists of 128-electrode EEG signals that were recorded at 5000 Hz in an electromagnetically shielded EEG lab specifically optimized to reduce environmental and electromagnetic noise. The resulting EEG data is therefore especially well suited for extracting information in higher frequencies\cite{Schirrmeister2017a}. The dataset contains recordings from 14 subjects with roughly 1000 trials each. Each trial has a duration of 4 s during which the subject was tasked to perform movements of either the left hand, right hand, both feet, or rest. We downsampled the data to 250 Hz to achieve faster training times  and common average re-reference the data (originally referenced to Cz) to obtain more easily interpretable visualizations.

\subsection{Network}
We use the network architecture proposed by Schirrmeister et al. (2017). They showed this architecture to perform well on task-related EEG data. The network details are shown in Figure \ref{fig:Architecture}. The first layer takes 128 channels with 522 time points as input. The trials are cropped into inputs of 522 time points using sliding windows with maximum overlap. We performed our analysis on the convolutional layers denoted by 1, 2, 3 and 4 in Figure \ref{fig:Architecture}. We omitted the first convolutional layer, because there are no non-linearities between its activations and the following convolutional layer which makes the activations of individual filters harder to interpret. Training was performed in the manner described in Schirrmeister et al. (2017) on each subject individually, resulting in 14 different models. Roughly ~800 trials per subject were used for training. Mean test accuracy over the 14 subjects was 88.6\% (std: 7.72).

\subsection{Signal perturbation}
\subsubsection{Amplitude perturbation}
To detect the phase-invariant response of filters to changes in the amplitude of specific frequencies, we used the perturbation correlation approach described in Schirrmeister et al. (2017). We adapted it by using multiplicative instead of additive noise to obtain perturbations with similar relative strength across the different frequencies. In case a filter extracts the amplitude of a certain frequency, a perturbation of the amplitude of that frequency should evoke a consistent change of activity in all units of that filter. An amplitude increase should evoke either an activation increase or an activation decrease in all units. The opposite should happen for an amplitude decrease.

Perturbation correlations were calculated for each layer and subject individually as following (\textbf{bold font} indicates differences to Schirrmeister et al. 2017).

\begin{enumerate}
\item Each trial $X_i$ in the training set of a subject was Fourier transformed to the frequency domain, resulting in amplitudes $A_\xi(X_{c,i})$ and phases $\theta_\xi(X_{c,i})$ for frequency $\xi$, trial $i$ and channel $c$.
\item \textbf{Independent gaussian-distributed perturbation factors $p^{A}_{\xi,c,i}{\sim}N(1,0.02)$ were multiplied with the amplitudes for each trial, channel and frequency, resulting in perturbed amplitudes $A^{p}_\xi(X_{i,c})=p^{A}_{\xi,i,c}*A_\xi(X_{c,i})$.}
\item Perturbed trials $X^{A}$ were reconstructed by inverse Fourier transform using the original phases and the perturbed amplitudes.
\item Activities $y_{f,i,j}$ for filters $f$ were calculated for the unperturbed trials and activities  $y^{A}_{f,i,j}$ were calculated for the perturbed trials
\item The mean difference of unit activation in a filter was calculated: $\bar{\Delta{y_{f,i}}}=\frac{1}{N_j}\sum_{j}^{ }(y_{f,i,j}-y^{A}_{f,i,j})$, where $N_j$ is the number of units in a filter.
\item Correlating the perturbations with the mean activation differences resulted in the amplitude perturbation correlation $\rho_{p_{\xi,c}^A,\bar{\Delta{y_{f}}}}=corr(p^{A}_{\xi,c},\bar{\Delta{y_{f}}})$.
\item \textbf{The absolute perturbation correlations were averaged over filters, channels, subjects, and several repetitions of the procedure resulting in the mean absolute amplitude perturbation correlation $\varrho^A_{l,\xi}$}
\end{enumerate}

The mean absolute amplitude perturbation correlation $\varrho^A_{l,\xi}$ should help understand the general behavior of individual layers in response to amplitude perturbations.

\subsubsection{Phase perturbation}
The response of filters to changes in the phase of certain frequencies was calculated similarly to the amplitude perturbation correlations. However, because of the cyclic nature of phase features, the change of activations in a filter resulting from a phase shift can not be quantified using the mean activation difference. Instead of taking the difference between original and perturbation activations, we calculated the correlation between both of them. In case of a filter that is sensitive to a specific phase in a specific frequency, a phase shift in that frequency should evoke a temporal shift in the unit activations of that filter, corresponding to the phase shift. Units of filters whose receptive field contained its specific phase in the original signal should activate less and units whose receptive field contains the specific phase in the perturbed signal should then activate more. Therefore, the original activations and the activations on the perturbed input should have a decreased correlation (less than 1). Activation and correlation should remain similar for phase-insensitive filters.

Additionally, we wanted to study only the effect of changing the overall phase of the signal, independent of the effect of increased or decreased phase synchronicity across EEG channels. To this aim, we did not perturb the phase in channels individually, but applied one phase perturbation of a certain frequency in all channels equally.

Phase perturbations were sampled from $p^{P}_{\xi,i}{\sim}N(0,\pi)$. Perturbed phases were calculated by shifting the phase: $P^{P}_\xi(X_i)=p^{P}_{\xi,i}+P_\xi(X_i)$. Perturbed signals $X^{P}$ were reconstructed by inverse Fourier transformation. The correlation between original and perturbation filter activations of a filter $f$ from trial $i$ is denoted by $\rho_{y_{f,i},y^{P}_{f,i}}=corr(y_{f,i},y^{P}_{f,i})$. Correlations between phase perturbations $p^{P}_{\xi}$ and filter activity correlations $\rho_{y_{f},y^{P}_{f}}$ were calculated identically to amplitude perturbations. The resulting mean absolute phase perturbation correlations for each layer is denoted as $\varrho^P_{l,\xi}$.

\subsection{Most-activating input windows}
In addition to examining how the individual layers respond to frequency-specific phase and amplitude, we were also interested in other characteristic features that might be learned by filters. However, direct interpretation of learned ConvNet filter weights is not trivial. Weights of a discriminatively trained model can both reflect what the model learned about the relation between the input signal and the classes, or what the model learned about class-independent noise covariances in the input \cite{Haufe2014}. Additionally, because of the hierarchical structure of the network, filters learned in later layers are not applied directly on the input, but applied on representations from previous layers.

We therefore decided to use the common approach to investigate the input windows that evoked the highest activations in a filter\cite{Girshick2013}. For each filter, we determined the 10\% highest unit activations over all training trials, while enforcing that each trial contributes at most one of those highest unit activations. For each of these highest unit activations, we determined its receptive field in the input signal, which we call the input window. Correspondingly, we call the input windows of the 10\% highest unit activations the most-activating input windows of a filter. We visualize the EEG signals in those input windows and their median to get an impression of the signal characteristics \textit{a filter is sensitive to}.

\subsubsection{Sine wave fitting}
We were further interested to test if filters were sensitive to a particular phase in the input signal and investigated if the learned features resembled parts of or complete sinusoidal curves. To quantify this, we standardized the individual most-activating input windows of each filter and performed a least squares fit of $y=o+a*cos(\xi*x+\theta)$ on the resulting standard scores. We also performed a simple linear fit of $y=m*x+b$ and compared the mean squared errors (MSEs) of the sinusoidal and the linear fits. The sinusoidal MSEs should be lower than the linear MSEs for filters sensitive to input windows resembling a sinusoidal in a specific phase.

Additionally, we also fitted the medians of most-activating input windows. The median of most-activating input windows should already resemble a sinusoid of a certain frequency if the input windows share a sinusoid of similar phase at that frequency. Additionally, we were interested to see which frequencies were most often used for the sinusoidal fits in each layer.
    \begin{figure}[t!]
        \includegraphics[width=0.48\textwidth]{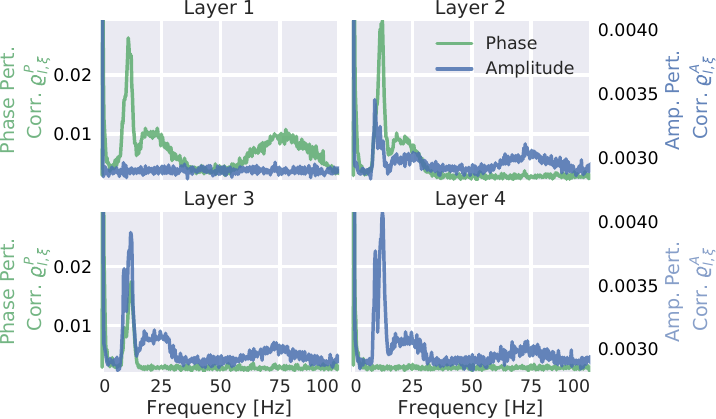}
        \caption{Mean of absolute phase and amplitude perturbation correlations for individual frequencies. The two correlation types have different scales, denoted by the left and right y-axis. As in Figure \ref{fig:PhaseAmpDev}, a clearly inverse relation between amplitude and phase correlations is visible. This visualization additionally showed that alpha (7-13 Hz), beta (13-30 Hz), and high gamma (50-100 Hz) frequency ranges each have a specific layer in which their phase correlation vanishes and their amplitude correlation saturates (high gamma in layer 2, beta in layer 3, and alpha in layer 4).}
        \label{fig:PhaseAmpSina}
    \end{figure}
\section{Results}
\subsection{Phase and amplitude sensitive layers}
The perturbation analysis showed that the earlier layers represent  more phase-specific features than later layers, while the later layers represent more phase-invariant amplitude features than the early layers. Figure \ref{fig:PhaseAmpDev} shows the average absolute phase perturbation correlation $\varrho^P_{l,\xi}$ and amplitude perturbation correlation $\varrho^A_{l,\xi}$ over the 4 convolutional layers. The figure shows a clearly opposing development of their respective average values across layers. $\varrho^P_{l,\xi}$ decreased and $\varrho^A_{l,\xi}$ increased with increasing layer depth, showing an almost exactly inverse relation between $\varrho^P_{l,\xi}$ and $\varrho^A_{l,\xi}$. Both correlations decrease/increase heavily from first to second layer and less in the following layers.
\setcounter{figure}{3}
\begin{figure}[h]
    \centering
    \includegraphics[width=0.48\textwidth]{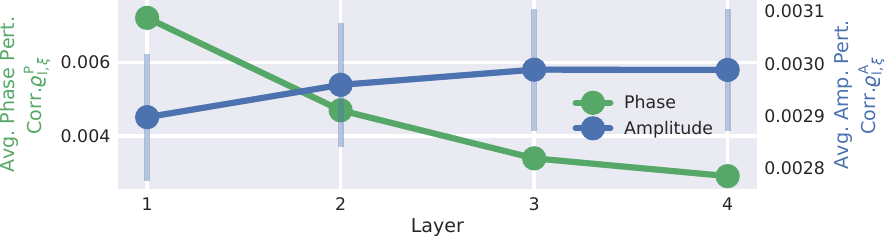}
    \caption{Mean phase and amplitude perturbation correlations over layers. Curves show mean perturbation correlation over all frequencies for each layer. Scales are different and written on the left and right y-axes. The error bars show the standard error over the subjects. Standard errors for phase and amplitude are similar, but much higher relatively in the amplitude correlation scale and therefore only visible there. In their respective scale, curves show a clearly inverse behavior over the layers with increasing amplitude correlations and decreasing phase correlations.}
    \label{fig:PhaseAmpDev}
\end{figure}

The perturbation correlations for individual frequencies showed a strong phase perturbation correlation $\varrho^P_{l,\xi}$ in the earlier layers 1 and 2 to phases in the alpha, beta, and high gamma range (Figure \ref{fig:PhaseAmpSina}). The overall phase perturbation correlation was highest in layer 1 and gradually became lower over layers 2, 3, and 4. Interestingly, for each frequency band (alpha, beta, and high gamma), there was one specific layer in which the phase perturbation correlations vanished completely. Phase perturbation correlations for high gamma vanished in layer 2, correlations for beta vanished in layer 3, and correlations for alpha vanished in layer 4. This vanishing of phase correlations for high gamma in layer 2 could be observed in all subjects. For 4 subjects, alpha did already vanish together with beta in layer 3.

An opposite behavior could be observed for amplitude correlations  $\varrho^A_{l,\xi}$ (Figure \ref{fig:PhaseAmpSina}). Notable phase insensitive amplitude perturbation correlations are only emerging in layer 2. Amplitude correlations of individual frequency bands peaked and saturated in the same layers in which the phase correlation vanished: High gamma amplitude perturbation correlation saturated in layer 2, beta in layer 3, and alpha in layer 4.
\setcounter{figure}{2}
    \begin{figure}[t!]
        \includegraphics[width=0.48\textwidth]{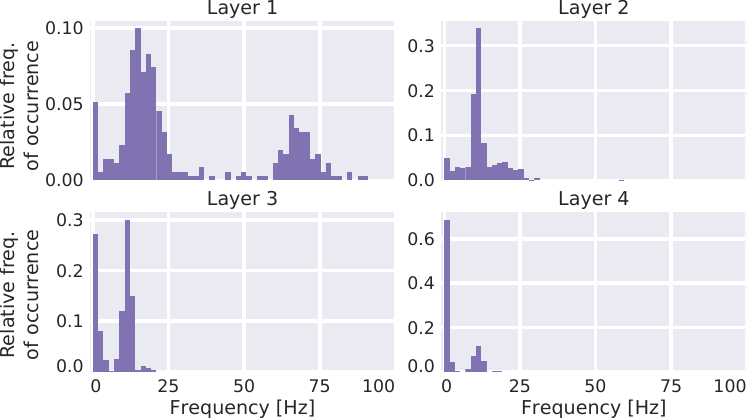}
        \caption{Relative frequency of occurrence for individual frequencies of the fitted sinusoids. Histograms show how often frequencies in that bin were used for a sinusoidal fit on medians in a layer.}
        \label{fig:PhaseAmpSinb}
    \end{figure}

\subsection{Sine wave fits to most-activating window EEG signals}
The mean squared error (MSE) of the most-activating input window sinusoidal and linear fits is shown in Figure \ref{fig:errors}. While still relatively large, the MSE of the sinusoidal fit was consistently lower than that of the linear fit. This means that the most-activating input windows could be better approximated by a sinusoid in all layers. The MSE for the sinusoidal fit for most-activating input windows was lowest in layers 1 and 2. The MSE of the sinusoidal fit approached the MSE of the linear fit for last layer.
\setcounter{figure}{4}
\begin{figure}[t]
    \centering
    \includegraphics[width=0.48\textwidth]{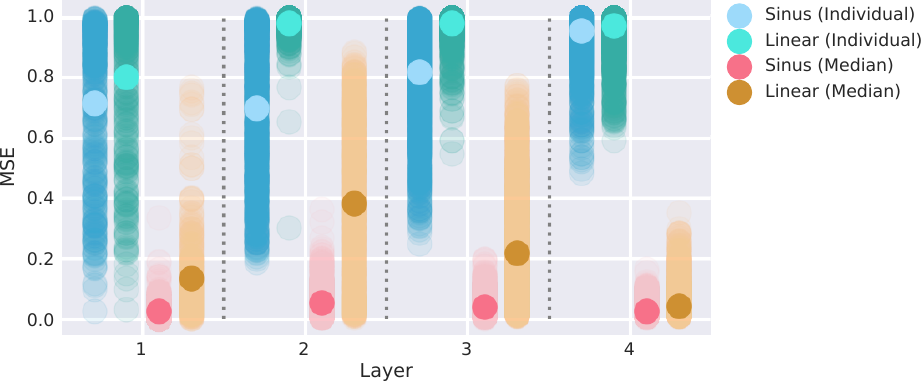}
    \caption{MSE of sinusoidal and linear fits on the most-activating input windows. Shown are distributions of mean squared errors and their mean for each layer, once fitted on the data in the individual input windows, and once fitted on the median across them. These distributions differed statistically significantly between sinusoidal and linear fits for all layers, both for fits to individual signals and for fits to median signals (Wilcoxon signed-rank test, all p-values below 1e-10).}
    \label{fig:errors}
\end{figure}

The MSE for sinusoids fitted on the median of the most-activating windows is also shown in Figure \ref{fig:errors}. The medians in all layers were again consistently better approximated by a sinusoidal fit compared to a linear fit. The MSE of the linear fits with the median was especially high for layers 2 and 3, but relatively low in layer 1. The sinusoidal MSE was low in all layers. Also, the distributions of the individual squared errors from sinusoidal fits were narrow for all layers. Similarly to the fit on individual input windows, the MSE of the sinusoidal fit approached the MSE for the linear fit in the last layer.

\begin{figure}[t]
    \centering
    \includegraphics[width=0.48\textwidth]{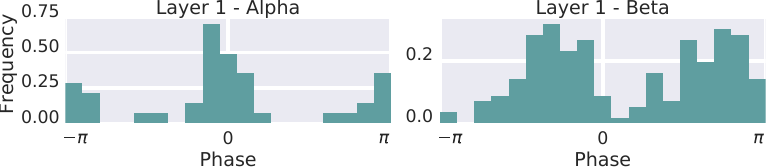}
    \caption{Histogram of phases of the fitted sinusoids. Sinusoids were fitted to the  medians of the most activating input windows per filter. Histograms are shown for phases of those sinusoidal fits that had a frequency in the alpha or beta band for layer 1 over all subjects. A clear bimodal distribution is visible on both frequency ranges.}
    \label{fig:PhaseHist}
\end{figure}
The distributions of the frequencies of the fitted sinusoids strongly resembled the phase correlation plots from the previous section (Figure \ref{fig:PhaseAmpSinb}). In layer 1, frequencies of the fitted sinusoids were in the alpha, beta, and high gamma bands. High gamma band frequencies disappear in layer 2, beta band frequencies in layer 3, and alpha frequencies are strongly reduced in the step from layer 3 to 4.

Furthermore, the distributions of phases of sinusoids fitted to medians with a fitted frequency in the alpha or beta band for layer 1 clearly showed bimodal distributions with 2 opposing peaks shifted by $\pi$ (Figure \ref{fig:PhaseHist}). However, distributions for other frequency bands or layers were not as clearly bimodal, but resembled more uniform distributions.

\subsection{EEG patterns in most-activating input windows}
\begin{figure}[t]
    \centering
    \begin{subfigure}[b]{0.23\textwidth}
        \includegraphics[width=\textwidth]{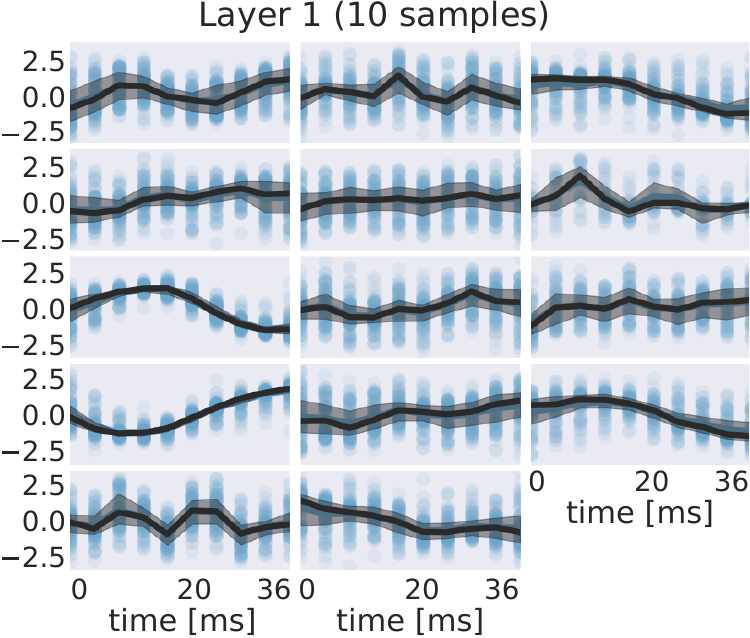}
        \label{fig:inp1}
    \end{subfigure}
    ~ %add desired spacing between images, e. g. ~, \quad, \qquad, \hfill etc. 
    \begin{subfigure}[b]{0.23\textwidth}
        \includegraphics[width=\textwidth]{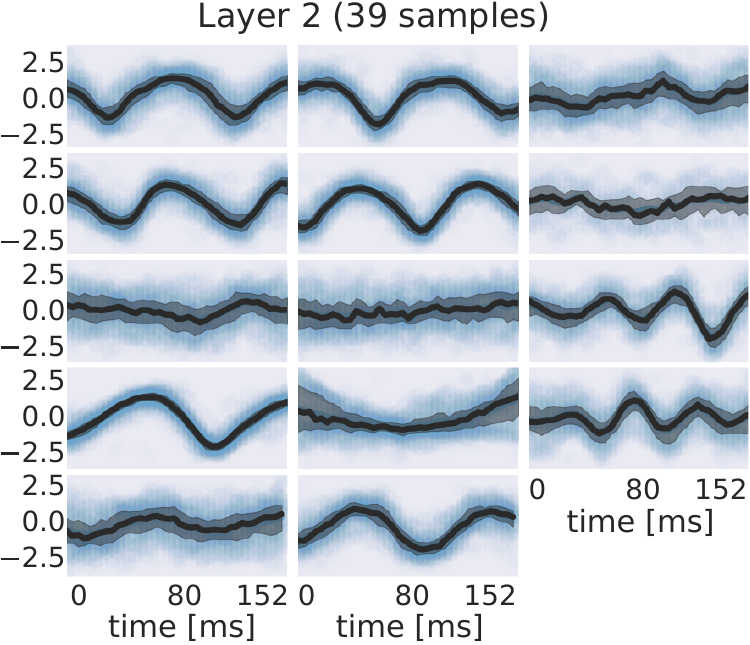}
        \label{fig:inp2}
    \end{subfigure}
    %\hfill
    \vspace{-10pt}
    \begin{subfigure}[b]{0.23\textwidth}
        \includegraphics[width=\textwidth]{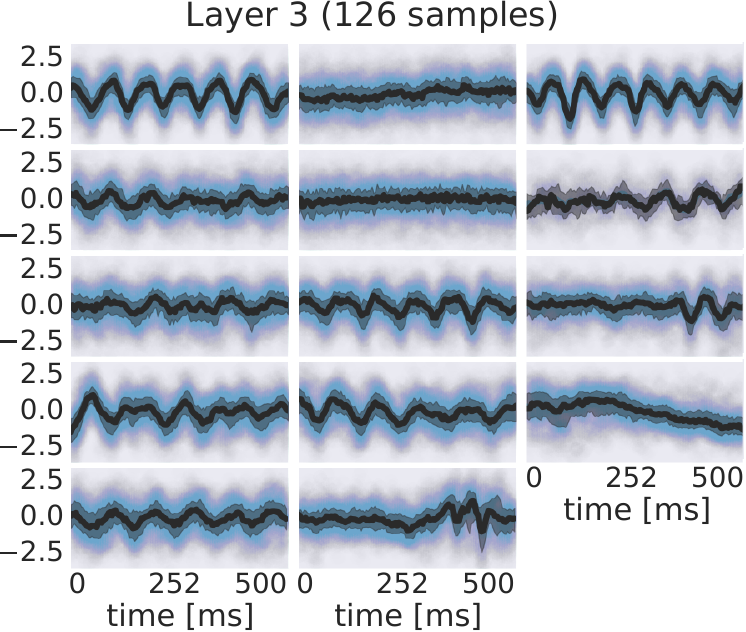}
        \label{fig:inp3}
    \end{subfigure}
    ~ %add desired spacing between images, e. g. ~, \quad, \qquad, \hfill etc. 
    \begin{subfigure}[b]{0.23\textwidth}
        \includegraphics[width=\textwidth]{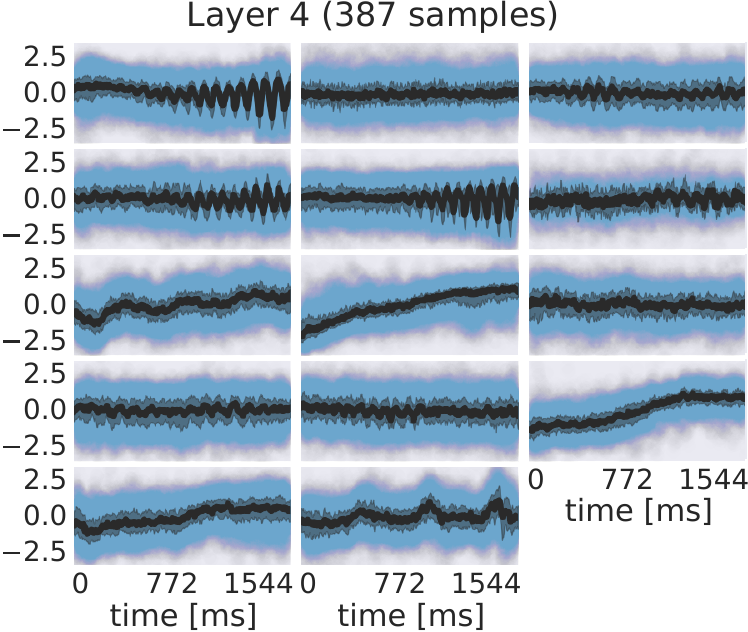}
        \label{fig:inp4}
    \end{subfigure}
    \caption{EEG signals in representative most-activating input window of one randomly sampled filter from each subject for each layer. Blue points are the standard scores of all most-activating input windows for a filter. Their median is shown in black and the interquartile range as a gray shaded area. Medians with respect to earlier layers often resemble parts of or complete sinusoids while medians in later layers resemble more complex patterns.}
    \label{fig:windows}
\vspace{-15pt}
\end{figure}

To investigate other features than frequency-specific phase or amplitude of sinusoidal signals, we visually inspected the most-activating input windows of a filter and their median values for each timepoint. Figure \ref{fig:windows} shows the most-activating input windows of one randomly sampled filter for each subject and layer. We show each such set of most activating input windows here at a representative electrode.

For several filters, a clearly defined structure was present in the median. The median plots for layer 1 show several examples of sinusoidal shapes in different frequencies. Medians of higher frequencies were sharper and the variance across input windows is larger. Medians of layer 2 revealed several smooth alpha sinusoids, but also examples of beta waves. In addition to that, there were some flat medians without an easily interpretable periodicity or other temporal structure. In layer 3, there were mostly examples for alpha waves. For some medians in layer 4, more complex temporal patterns emerged. Those medians were relatively flat at the beginning of the input windows, but showed an oscillatory pattern with increasing amplitude in the later parts of the input windows. Also, the oscillatory patterns in some examples resembled mu waves more closely than pure sinusoids. Such patterns were found in several subjects.
\section{Discussion}
The findings of this study provide insight about the internal representation of EEG in a ConvNet trained for EEG decoding and how these representations develop over the different convolutional layers. Schirrmeister et al. (2017) showed correlation of spectral amplitude perturbations with activation changes for the final classification layer. We expanded this approach by investigating the influence of both spectral amplitude and phase through the sequence of intermediate layers. We showed that filters of later layers of the ConvNet extract information about the spectral amplitudes of a signal by combining information from filters that detect phase-specific periodic, often sinosoidal patterns of a certain frequency in early layers. Spectral amplitude is highly informative in our case of trial-wise motor decoding, whereas spectral phase is more informative in continuous motor behavior decoding\cite{Hammer2013}.

Our visualizations revealed a relation between typical EEG frequency bands and the intermediate representations of the trained ConvNets. Different intermediate layers showed a specialization to either the alpha, beta, or high gamma band. In our data, information about the EEG amplitude in higher frequency bands was extracted early in the network and then forwarded through later layers. Additionally, we observed bimodal distributions of phases from fitted sinusoids having a distance of $\pi$ between the peaks. This is a notable difference to the bases of a conventional Fourier transformation, where the output coefficients for one frequency are defined as the dot product of the signal with a sinusoid of that frequency and the sinusoid shifted by $\frac{\pi}{2}$ (i.e., the cosine and sine function for that frequency). This suggests networks are not using exactly the same bases as the Fourier transformation. As the size of the temporal receptive fields grows from early to later layers (cf. Schirrmeister et al. 2017), the spectral decomposition learned by the networks might have higher time resolution for higher frequencies (here: high-gamma components), and lower time resolution for lower frequency bands (here: alpha and beta bands), similar to Wavelet spectral analyses. Further investigations would however be necessary to clarify how the properties of the spectral decomposition learned by ConvNets on EEG relates to classical spectral estimation techniques.

In addition to a clear sensitivity of filters to basic spectral features, the visual analysis of the average (median, see Figure \ref{fig:windows}) EEG data in the most-activating input windows showed a preference of more complex features in some filters of the last layer. Thus, filters of later layers in networks trained for EEG decoding appear to not only learn to detect basic spectral phase and amplitude features, but also more complex combinations of those. Filters in later layers could be analyzed to reveal more of such complex features and thereby possibly increase our understanding of the underlying signals. This would be similar to networks trained for image classification, for which filters of later layers have been shown to detect complex structures like faces, parts of objects or abstract categorical features\cite{Zeiler2013}.

\section{Conclusion \& Outlook}
This study provided insights into the internal representation of spectral features of convolutional networks trained for EEG decoding. We described possible mechanisms of how ConvNets learn to hierarchically combine low- to higher-level EEG features through the sequence of convolutional layers of the network.

As future steps, it would be interesting to examine how these representations of spectral features depend on the network architecture or decoding task. Regarding the network architecture, one could investigate if one convolutional layer per physiologically meaningful EEG frequency band is a natural choice resulting in potential performance or interpretability advantages. Regarding the decoding task, one could train the network to decode continuous movement parameters for which EEG phase may play a more important role than in the classification task examined here (cf. \cite{Hammer2013}). Furthermore, examining how the internal spectral representations of ConvNets relate to classical spectral estimation techniques for EEG analysis might possibly provide a new alternative to classical spectral analyses. Progress along these lines may help optimizing ConvNets for EEG decoding and visualization, and in exploring more complex EEG features.

%\bibliographystyle{ieeetr}
%\bibliography{bib}

\end{document}